# A Deeper Look into Hybrid Images

By

**Jimut Bahan Pal**
**B1930050**
Semester - I

Under the guidance
of

**Tamal Maharaj**

Submitted to the Department of Computer Science in
partial fulfilment of the requirements
for the degree of M.Sc.

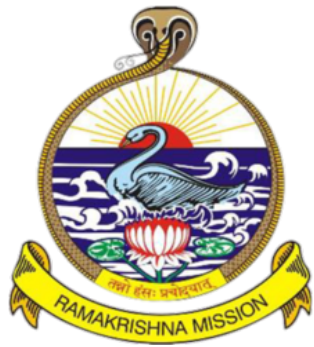

**Ramakrishna Mission Vivekananda Educational and Research Institute**
**Howrah - 711202**
**January, 2020**

# CANDIDATES' DECLARATION

This is to certify that the work presented in this thesis, titled, "A Deeper Look into Hybrid Images", is the outcome of the investigation and research carried out by me under the supervision of Tamal Maharaj.

It is also declared that neither this thesis nor any part thereof has been submitted anywhere else for the award of any degree, diploma or other qualifications.

---

Jimut Bahan Pal
B1930050

# CERTIFICATION

This thesis titled, **"A Deeper Look into Hybrid Images"**, submitted by Jimut Bahan Pal as mentioned below has been accepted as satisfactory in partial fulfillment of the requirements for the degree M.Sc. in Computer Science in January, 2020.

**Group Members:**

    **Jimut Bahan Pal**
    jimutbahanpal@yahoo.com

**Supervisor:**

    ______________________________
    Tamal Maharaj
    Ph.D. from University at Buffalo NY USA
    Department of Computer Science
    Ramakrishna Mission Vivekananda Educational and Research Institute

# ACKNOWLEDGEMENT

It is ritual that scholars express their gratitude to their supervisors. This acknowledgement is very special to me to express my deepest sense of gratitude and pay respect to my supervisor, Tamal Maharaj, Department of Computer Science, for his constant encouragement, guidance, supervision, and support throughout the completion of my project. His close scrutiny, constructive criticism, and intellectual insight have immensely helped me in every stage of my work.

Im grateful to my father, Dr. Jadab Kumar Pal, Deputy Chief Executive, Indian Statistical Institute, Kolkata for constantly motivating and supporting me to develop this documentation along with the application. Finally, I acknowledge the help received from all my friends and well-wishers whose constant motivation has promoted the completion of this project.

Kolkata                                                                                           Jimut Bahan Pal
January, 2020

# Contents

# List of Figures

# ABSTRACT

*Hybrid images* was first introduced by Olivia et al., that produced static images with two interpretations such that the images changes as a function of viewing distance. Hybrid images are built by studying human processing of multiscale images and are motivated by masking studies in visual perception. The first introduction of hybrid images showed that two images can be blend together with a high pass filter and a low pass filter in such a way that when the blended image is viewed from a distance, the high pass filter fades away and the low pass filter becomes prominent. Our main aim here is to study and review the original paper by changing and tweaking certain parameters to see how they affect the quality of the blended image produced. We have used exhaustively different set of images and filters to see how they function and whether this can be used in a real time system or not.

# Chapter 1

# Introduction

## 1.1 What are hybrid images?

Hybrid images are a form of illusion which exploits the multiscale visual perspective of human vision [1]. It is formed by superimposing two images of different spatial scales, i.e., one by passing a high pass filter which captures the prominent version of the images and the other by passing a low pass filter. The image becomes a function of distance of vision, which means that when the image is viewed from close, it will capture the prominent version of the image and when it is viewed from distance, it will capture the lowpass version of the image, i.e., the blurry image.

## 1.2 Creating hybrid images

Here we can see that Figure 1.1, represents a combination of two images. The image above is of a fish, and the image below, is of a plane. When the image of a fish is passed through a high pass filter, it will capture the high pass version of the image, and when the image of the airplane is passed through the lowpass version of the filter, it will capture the blurry part of the image.

Both the images are added so that the addition of intensity values does not exceed the maximum intensity i.e., 255 and in most cases, they are added equally, i.e., 0.5 of the first image and 0.5 of the second image. When we see closely, we can see that the resultant image is comprised of a blended version of both the images, and the fish is prominent more in the blended version of the upper result, while the plane is more prominent in the blended version of the second image.

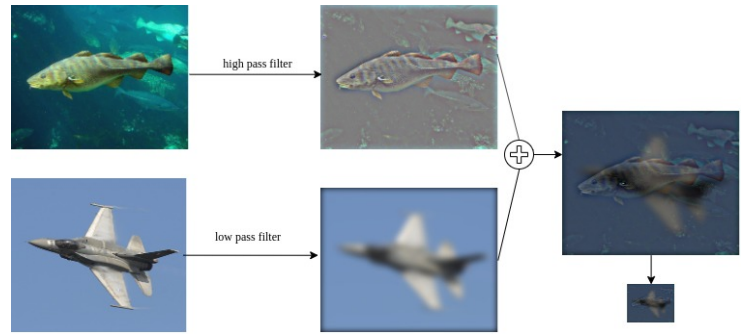

Figure 1.1: Blending of fish and plane resulting in hybrid image.

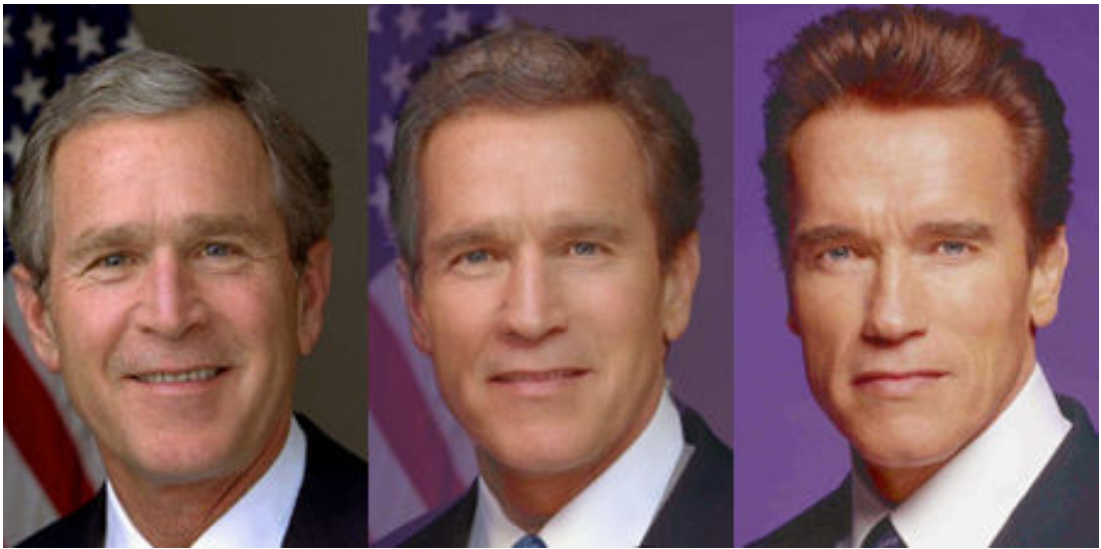

Figure 1.2: Three frames forming a morph from George Bush to Arnold Schwarzenegger (Source: Wikipedia).

## 1.3 Some other types of image blending

Another version of image blending is known as morphing. Here one image changes to another image by applying certain dissolving techniques as shown in Figure 1.2. Unlike hybrid images, which is a function of distance, i.e., **Image = f(distance)**, morphing is used extensively since the 90's for television advertisements. Here an image changes with time, i.e., **Image = f(time)**. It changes the shape of one image to another with seamless transition [2]. Hybrid images can be used in the field of multimedia when we need to morph a picture with the function of distance, i.e., a transition occurs when we move away or towards a certain picture. Unlike other techniques, this is a unique way of blending two images.

# Chapter 2

# Designing Hybrid Images

Let us consider two filters $G_1$ and $1 - G_2$, and two images $I_1$ and $I_2$. The filter $G_1$ is a low pass filter and is generally a Gaussian filter. Similarly, the filter $G_2$ is a high pass filter and is generally a Laplacian of Gaussian filter [3]. Laplacian filters are derivative filters which are used to find the areas of rapid change in images, i.e., it is used to find edges. Since derivative filters are very sensitive to noise [4], it is common to smooth the image (i.e., applying gaussian filter) before applying the Laplacian. This is a two step process called the Laplacian of Gaussian (LoG) operation.

## 2.1 The Gaussian filter

The impulsive response of this type of filter is a Gaussian function (and approximation to it, since true Gaussian is continuous and is physically unrealizable). The Gaussian function has the minimum possible group delay [5]. Mathematically, a Gaussian filter modifies the input signal with the Gaussian function via Weierstrass transformation.

One dimensional Gaussian filter has an impulse response given by,

$$g(x) = \sqrt{\frac{a}{\pi}} e^{-a.x^2} \tag{2.1}$$

In two dimensions, a Gaussian distribution is given by,

$$G_\sigma(x, y) = \frac{1}{2\pi\sigma^2} e^{-\frac{x^2+y^2}{2\sigma^2}} \tag{2.2}$$

Here, $x$ is the distance from the origin in the horizontal axis, $y$ is the distance from the origin in the vertical axis, and $\sigma$ the standard deviation of the Gaussian distribution.

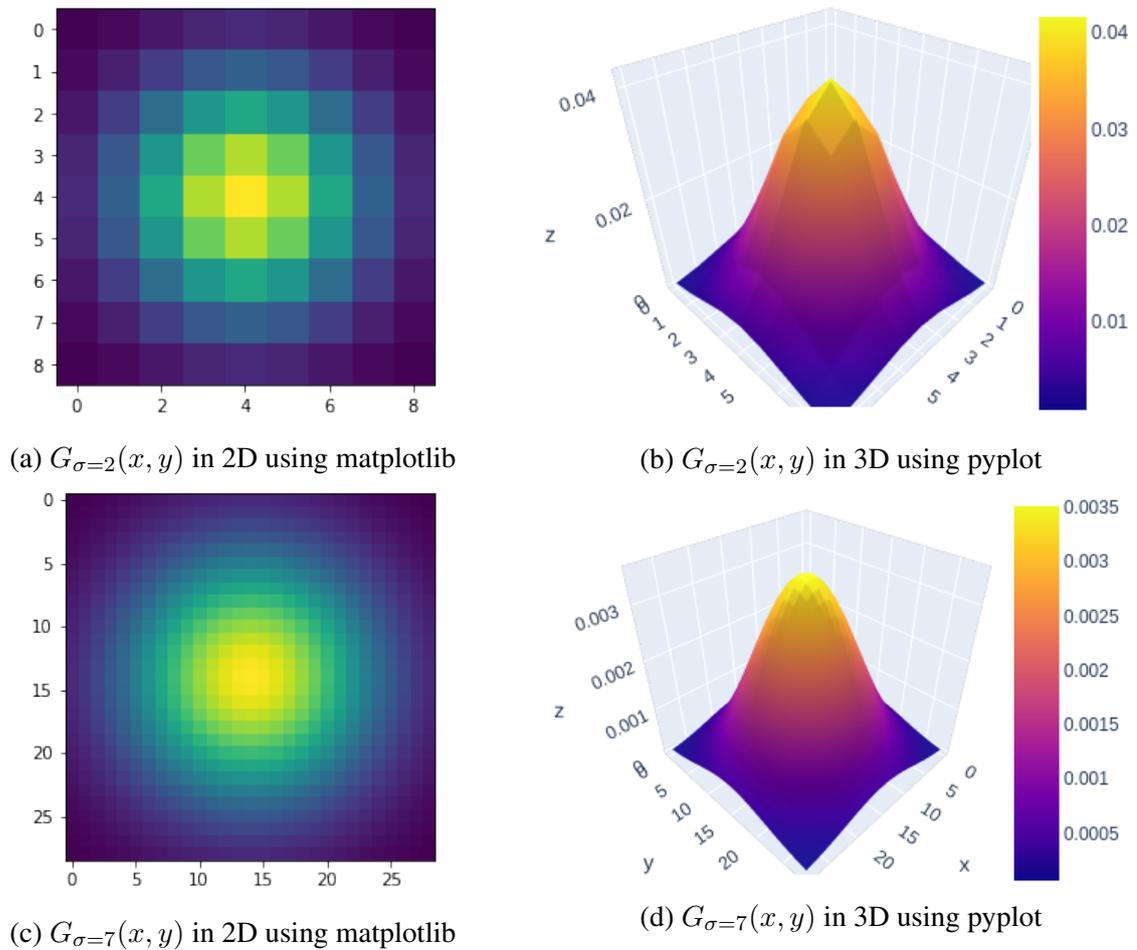

Figure 2.1: A figure containing the Gaussian filters of $\sigma = 2$ (top) and $\sigma = 7$ (bottom).

The Gaussian function is for x $\in (-\infty, \infty)$ theoretically, so it will need an infinite window length. Since the function decays quite fast, we don't need such huge window length, though it might introduce errors, but for different cases a different window function is used. A 3×3 Gaussian filter can be written as,

$$G_{3\times 3} = \frac{1}{16}\begin{bmatrix} 1 & 2 & 1 \\ 2 & 4 & 2 \\ 1 & 2 & 1 \end{bmatrix}$$

Here a discrete value for each element in window is used, and it is normalized by the $\Sigma_{(i,j)}$ which gives quite a good result. The $\Sigma_{(i,j)}$ is such that the total value of the intensity of a given pixel is $\leq 255$.

We can visualise the 2 dimensional and 3 dimensional structure of Gaussian filter in Figure 2.1. By observing the plot carefully, we can figure out that the 3-dimensional plot for the Gaussian for $\sigma = 2$ is rougher than the plot for $\sigma = 7$. The more the value of $\sigma$ the smoother the plot becomes.

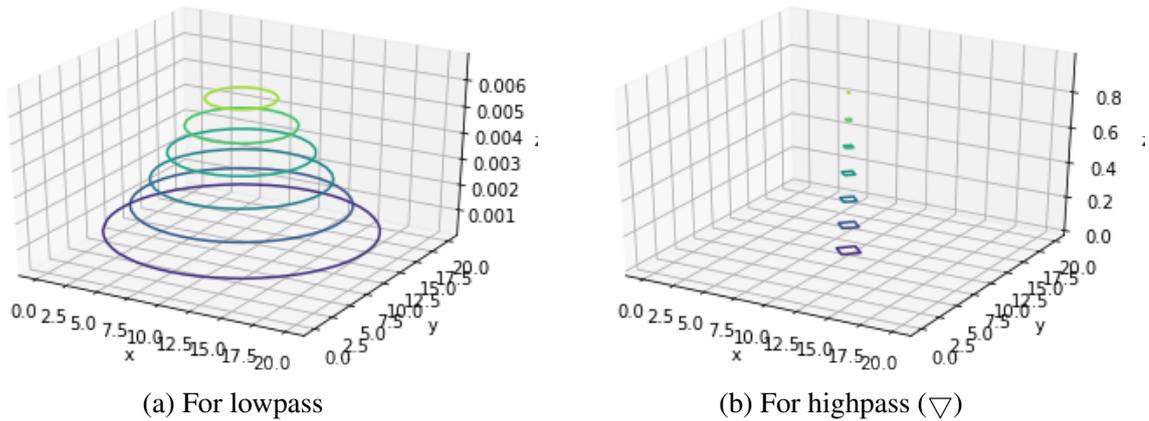

(a) For lowpass

(b) For highpass ($\triangledown$)

Figure 2.2: A figure containing the Contours for $G_{\sigma=5}(x,y)$ in 3D using matplotlib.

We can find the size of the filter by performing the following computations,

$$S = \sigma * 4 + 1 \tag{2.3}$$

Here, $S$ is the size of the filter and $\sigma$ is the standard deviation of the Gaussian filter.

If we think of the contours, the lowpass filter will have a smoother contour than a high pass filter. Here, Figure 2.2 shows the different contours of different filters. It is obvious the more the sigma, the more spread the contours is, i.e., the smoother the 3D plot is.

When the Gaussian filter is applied to the picture of Marylin as shown in Figure 2.3, with $\sigma = 2$, $\sigma = 4$, and $\sigma = 7$, the image becomes more blurry and the low pass component becomes visible from a distance.

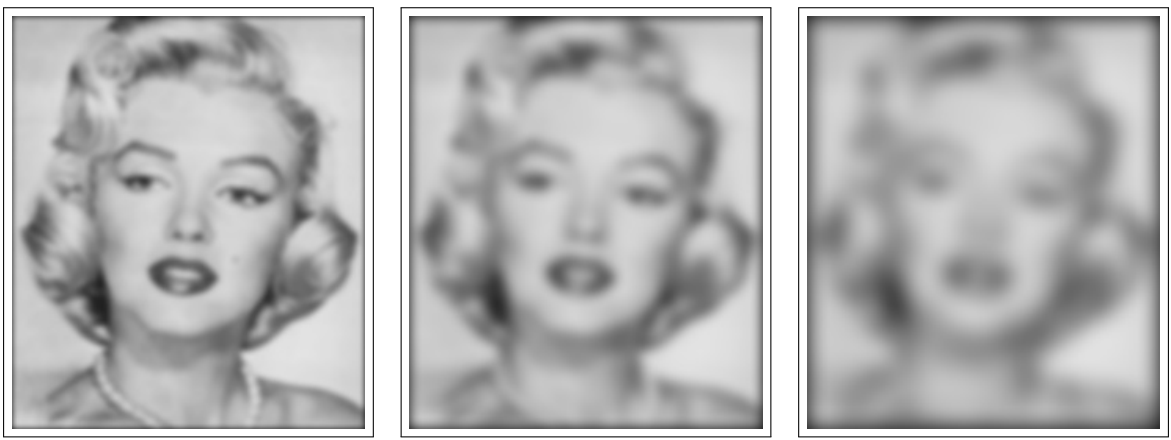

Figure 2.3: Applying lowpass filter to a picture of Marylin with $\sigma = 2$, $\sigma = 4$, and $\sigma = 7$.

## 2.2 The Laplacian filter

A Laplace operator may detect edges as well as noise, so it may be desirable to smooth the image first, to suppress the noise before using Laplace for edge detection [6].

$$\bigtriangledown \left[G_\sigma(x,y) * I(x,y)\right] = \left[\bigtriangledown G_\sigma(x,y)\right] * I(x,y) = LoG * I(x,y) \quad (2.4)$$

The first equal sign is due to the fact that,

$$\frac{d}{dt}[h(t) * f(t)] = \frac{d}{dt}\int f(\tau)h(t-\tau)d\tau = \int f(\tau)\frac{d}{dt}h(t-\tau)d\tau = f(t) * \frac{d}{dt}h(t) \quad (2.5)$$

For making the computation less, we can obtain the Laplacian of Gaussian $\bigtriangledown G_\sigma(x,y)$ first and then convolve it with the input image as shown below,

$$\frac{\partial}{\partial x}G_\sigma(x,y) = \frac{\partial}{\partial x}e^{\frac{-(x^2+y^2)}{2\sigma^2}} = -\frac{x}{\sigma^2}e^{\frac{-(x^2+y^2)}{2\sigma^2}} \quad (2.6)$$

and we again take a partial derivative to get (Note: we are omitting the normalizing coefficient $\frac{1}{\sqrt{2\pi\sigma^2}}$),

$$\frac{\partial^2}{\partial x^2}G_\sigma(x,y) = \frac{x^2}{\sigma^4}e^{\frac{-(x^2+y^2)}{2\sigma^2}} - \frac{1}{\sigma^2}e^{\frac{-(x^2+y^2)}{2\sigma^2}} = \frac{x^2-\sigma^2}{\sigma^4}e^{\frac{-(x^2+y^2)}{2\sigma^2}} \quad (2.7)$$

Similarly, by applying double partial derivatives to y we can get,

$$\frac{\partial^2}{\partial y^2}G_\sigma(x,y) = \frac{y^2-\sigma^2}{\sigma^4}e^{\frac{-(x^2+y^2)}{2\sigma^2}} \quad (2.8)$$

Now, we can define the convolution kernel as,

$$LoG \stackrel{\bigtriangledown}{=} \bigtriangledown G_\sigma(x,y) = \frac{\partial^2}{\partial x^2}G_\sigma(x,y) + \frac{\partial^2}{\partial y^2}G_\sigma(x,y) = \frac{x^2+y^2-2^2}{\sigma^4}e^{\frac{-(x^2+y^2)}{2\sigma^2}} \quad (2.9)$$

Hence the Gaussian $G_\sigma(x,y)$, $G'_\sigma(x,y)$ (its first derivative) and $\bigtriangledown G_\sigma(x,y)$ (its second derivative) can be shown in Figure 2.6

We can visualize the 2 dimensional and 3 dimensional structure for the Laplacian filters in Figure 2.4. By viewing the matrix for the filter, we can see that there is a sudden depression in the middle, this is actually the inverted kernel for the above figure. The main aim of the Laplacian filter is to capture the sudden change in intensity values which becomes prominent

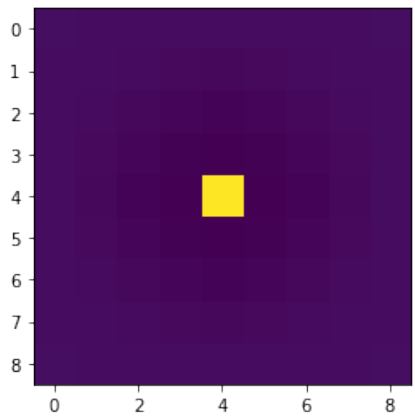
(a) $\nabla G_{\sigma=2}(x,y)$ in 2D using matplotlib

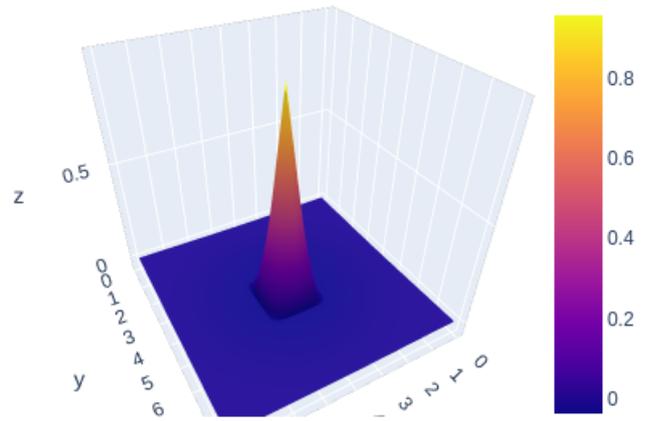
(b) $\nabla G_{\sigma=2}(x,y)$ in 3D using pyplot

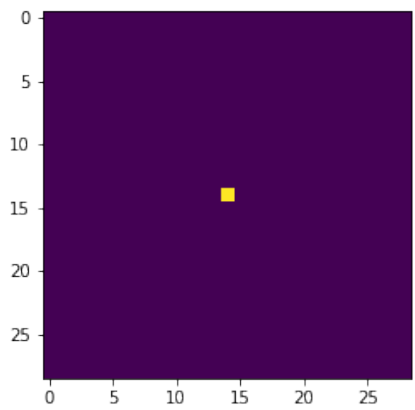
(c) $\nabla G_{\sigma=7}(x,y)$ in 2D using matplotlib

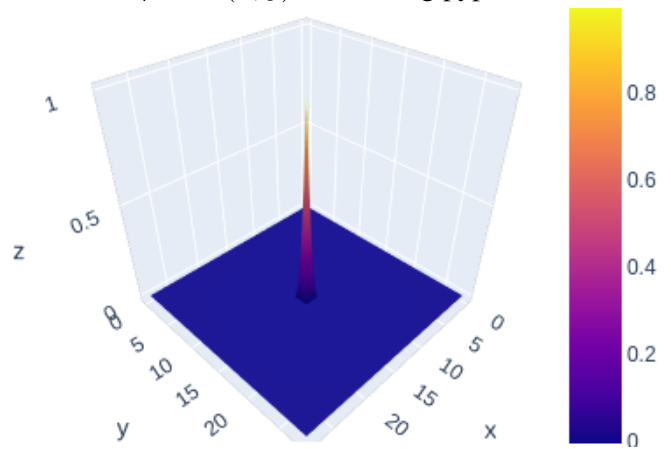
(d) $\nabla G_{\sigma=7}(x,y)$ in 3D using pyplot

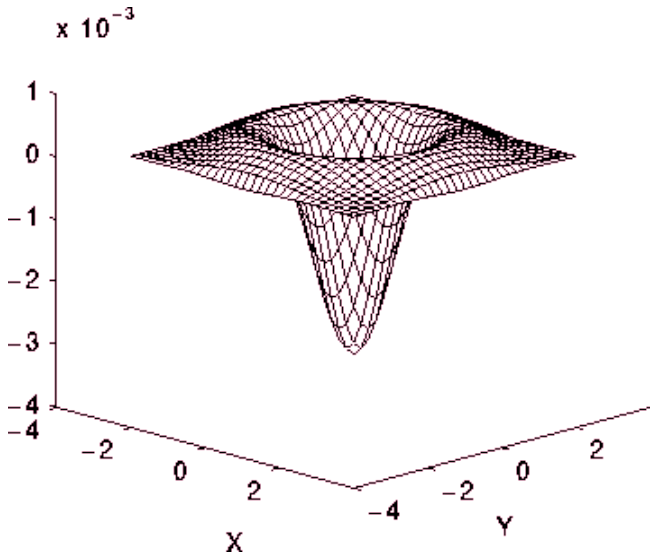
(e) A contour for inverted Laplacian filter (Source: UOED)

(f) A inverted Laplacian filter (Source: UOED)

Figure 2.4: A figure containing the high pass filters of $\sigma = 2$ (top), $\sigma = 7$(middle) and a Laplacian filter.

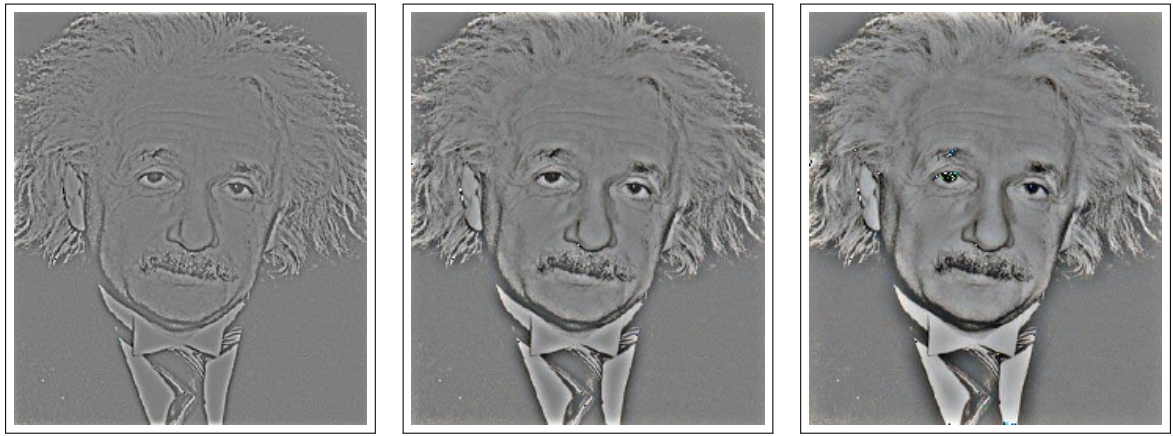

Figure 2.5: Applying highpass filter to a picture of Einstein with $\sigma = 2$, $\sigma = 4$, and $\sigma = 7$.

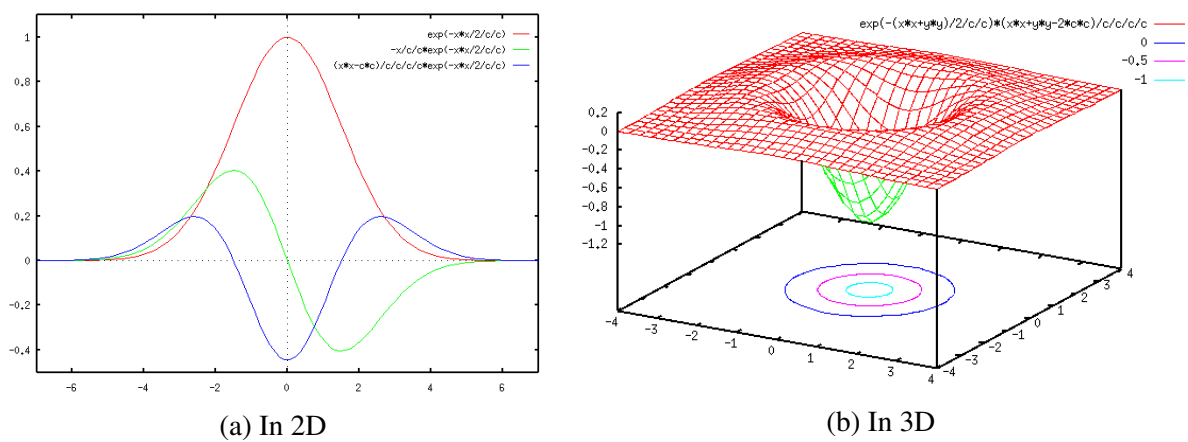

(a) In 2D  (b) In 3D

Figure 2.6: A figure containing the plots for $G_\sigma(x, y)$, $G'_\sigma(x, y)$ and $\triangledown G_\sigma(x, y)$ (Source: HMC).

when viewed from a closer distance.

When the Laplacian filter is applied to the picture of Einstein as shown in Figure 2.5, with $\sigma = 2$, $\sigma = 4$, and $\sigma = 7$, the image becomes more prominent and the high pass component becomes more visible from closer distance. Unlike Gaussian filter, the behaviour of the Laplacian filter is opposite, i.e., when the $\sigma$ increases, the filter captures the high frequency component from the image.

## 2.3 Theory of Convolutions

The convolution of two continuous signals [7] can be written as,

$$y(t) = h(t) * x(t) \stackrel{\triangledown}{=} \int_{-\infty}^{\infty} x(\tau)h(t-\tau)d\tau = \int_{-\infty}^{\infty} h(\tau)x(t-\tau)d\tau \qquad (2.10)$$

Convolutions are also associative in nature,

$$h * (g * x) = (h * g) * x \tag{2.11}$$

The application of the above equation can be appreciated if we consider $y(z)$ is the output of a system characterized by its impulse response function $h(t)$ with input $x(t)$. The convolution in discrete form which is actually used in the field of computer vision can be shown using the following equation,

$$y(n) = \sum_{m=-\infty}^{\infty} x(n-m)h(m) = \sum_{m=-\infty}^{\infty} h(n-m)x(m) \tag{2.12}$$

In general case, $h(m)$ is finite,

$$h(m) = \begin{cases} h(m) & |x| \leq k \\ 0 & |x| > k \end{cases}$$

The convolution becomes,

$$y(n) = \sum_{m=-k}^{k} x(n-m)h(m) \tag{2.13}$$

In almost all the cases, filter $h(m)$ is symmetric in image processing, i.e.,

$$h(-m) = h(m)$$

which forms the resulting equation,

$$y(n) = \sum_{m=-k}^{k} x(n+m)h(m) \tag{2.14}$$

## 2.4 Comparison of images formed by two of the filters

We will now compare the images formed by both the filters. The image of Einstein is passed through a high pass filter and a low pass filter as shown in Figure 2.7. The images when combined with a weighted value and normalized to 255 will produce a blending such that the low frequency component of the image is visible from a distance and the high frequency component is visible when looked closer to the image.

Let us now see the blending in action. We apply a filter of $\sigma = 7$ to the image of a cat (high

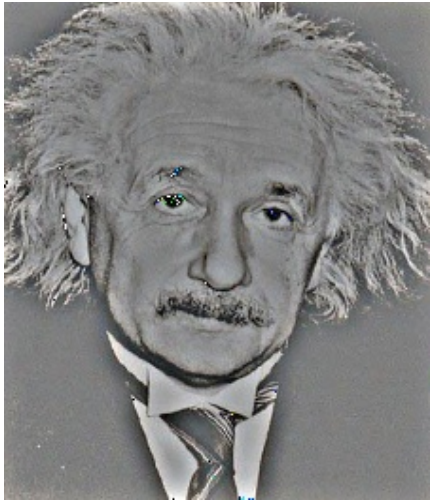
(a) By applying low pass filter

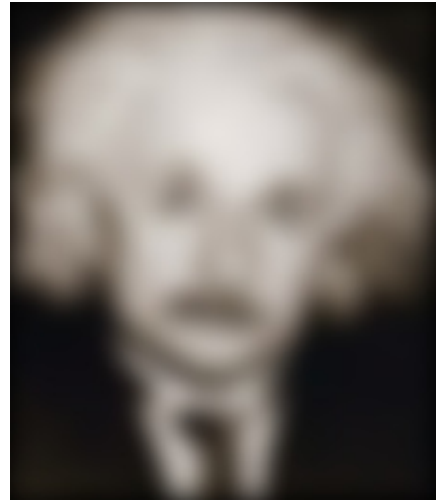
(b) By applying high pass filter

Figure 2.7: A figure containing the images of Einstein when applied a high pass and a low pass filter of $\sigma = 7$.

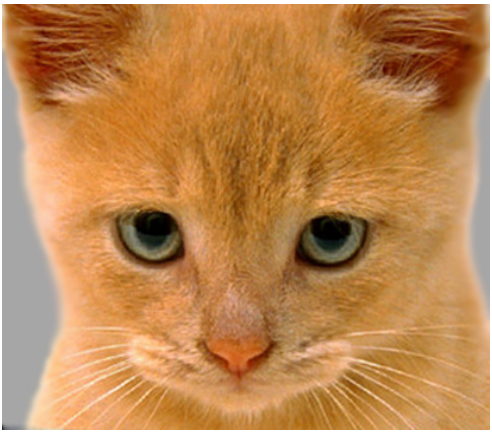
(a) The original image of the cat

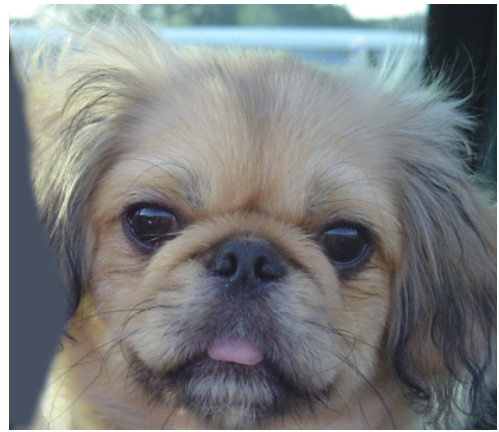
(b) The original image of the dog

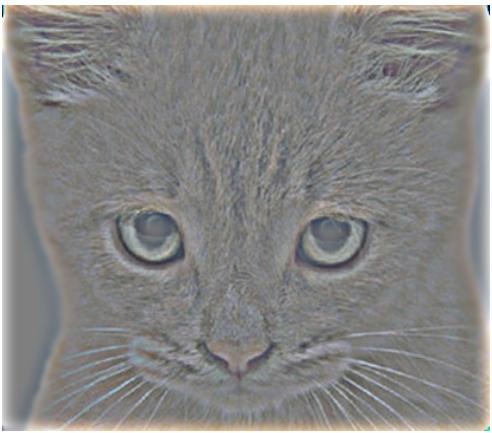
(c) By applying high pass filter to the cat image

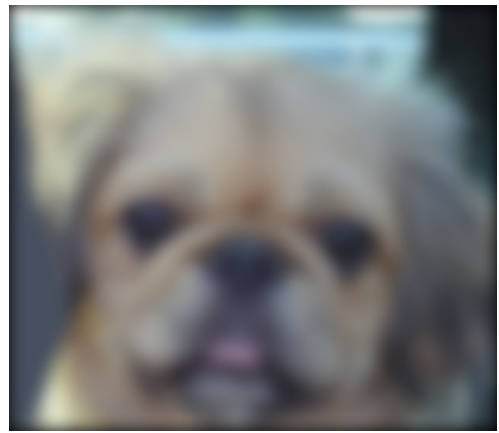
(d) By applying low pass filter to the dog image

Figure 2.8: A figure containing the images of cat and dogs when passed through a filter of $\sigma = 7$.

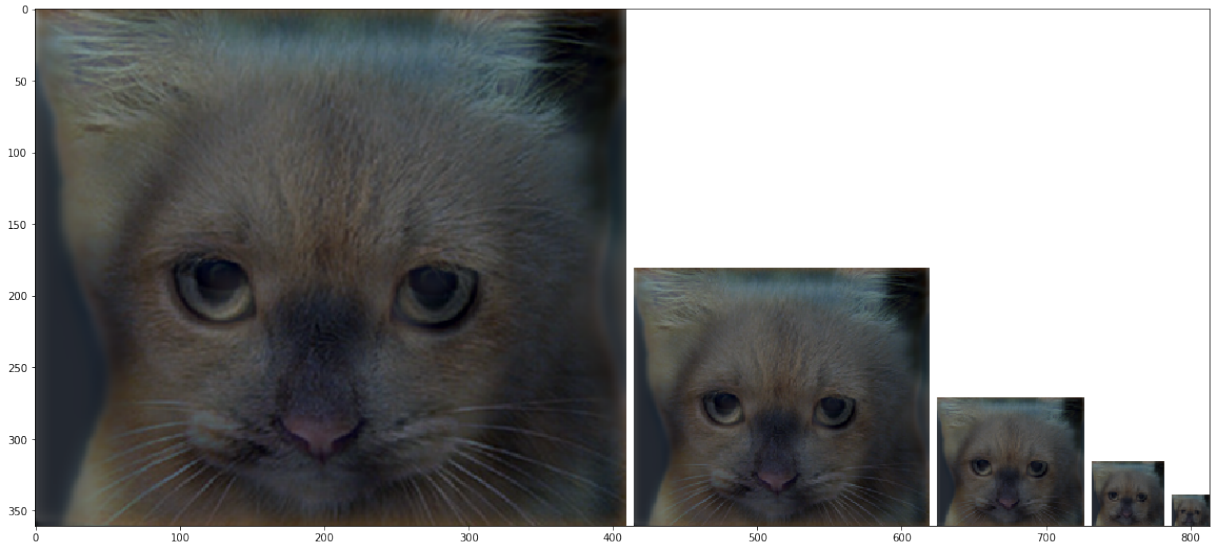

Figure 2.9: The blended image of cat and dog.

pass) and the image of a dog (low pass) as shown in Figure 2.8. The images extracted are then blended to form a blended image as shown in Figure 2.9. This is simple, and the images are free of noise, we can even tweak the parameter of weighted blending by hand and decide by how much one image should be blended to the other image. For making matters simple we just scale down the image so that it may appear that the image is moving away from the observer. By looking carefully, we will see that the image of the cat appears to be visible from the front, i.e., the first image of the figure. As the image goes away, i.e., the image becomes smaller and smaller we will find that the image of the dog becomes more prominent. This is exploiting the visual cortex system of a human which distinguishes an image formed by a high pass filter and an image formed by a low pass filter [8].

Similarly we apply a filter of $\sigma = 7$ to the image of a motorcycle (high pass) and the image of a bicycle (low pass) as shown in Figure 2.10. The images extracted are then blended to form a blended image as shown in Figure 2.11. The images given are free of noise and so it is not necessary to blur the image before applying the high pass filter. Here we can see that the blended image is formed by applying a low pass filter to the image of the bicycle which is more visible from a distance and a high pass filter to the image of motorcycle which is visible from a closer distance.

Finally we apply a filter of $\sigma = 7$ to the image of bird and plane and the resulting image is shown in Figure 2.12. We will now apply these techniques to images of different sizes in the next section.

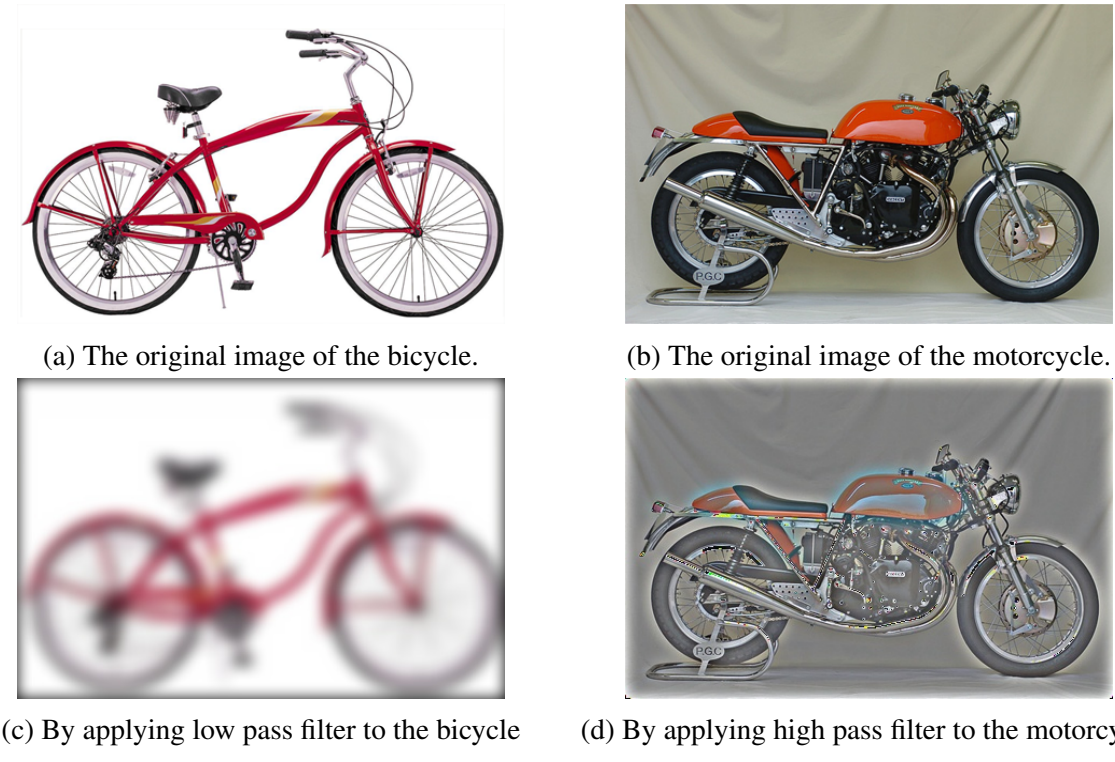

(a) The original image of the bicycle.  (b) The original image of the motorcycle.

(c) By applying low pass filter to the bicycle  (d) By applying high pass filter to the motorcycle

Figure 2.10: A figure containing the images of bicycle and motorcycle when passed through a filter of $\sigma = 7$.

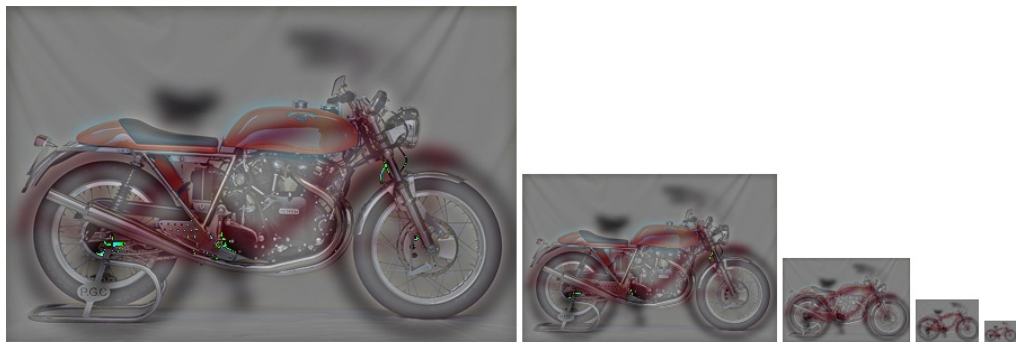

Figure 2.11: The blended image of bicycle and motorcycle.

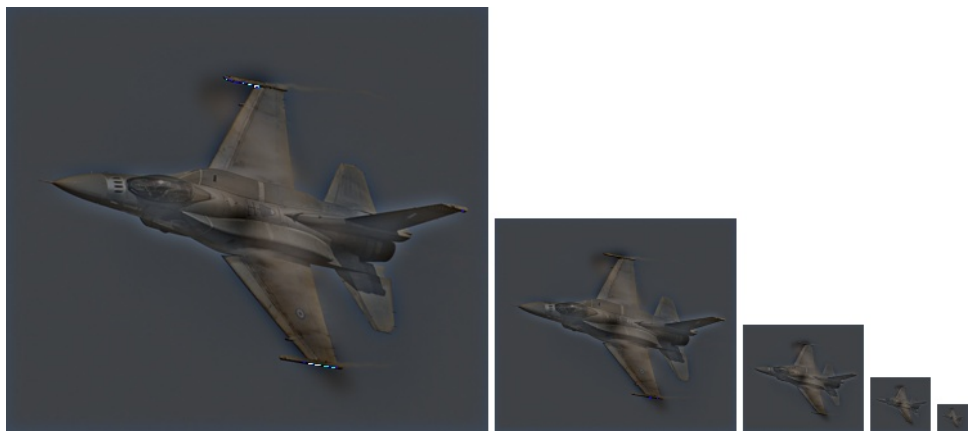

Figure 2.12: The blended image of a bird (low pass) and plane (high pass) by applying $\sigma = 7$ to both the images.

# Chapter 3

# Applying to real world examples

We have tested the algorithm to some toy examples [9]. We will now test this algorithm to some new pictures which are even larger than the examples tested before. We have noticed by trial and error that when the picture size increases we need to increase the cutoff frequency else it is not able to capture the more prominent version of the image, since in a large image the pixels are spread away more and by setting a small $\sigma$ we will not be able to capture a lot from the surroundings. We also notice that as the size of the image increases the amount of computation increases as there are three channels and the algorithm need to compute the filter to each of these channels.

## 3.1 Applying to images of Naruto and Sasuke

We first load the image of Naruto which has an original dimension of $1577 \times 894 \times 3$ as shown in Figure 3.1a. Then we load the image of Sasuke which have a dimension of $991 \times 721 \times 3$ as shown in Figure 3.1b.

We select the minimum of the dimension of two image and scale down the larger image, i.e., image of Naruto to the image of Sasuke. We do this because if the images are inconsistent in shape and sizes, then it becomes difficult to blend and add the two images. We apply a low pass filter of $\sigma$ = 30 to the image of Naruto as shown in Figure 3.2a and similarly we apply a high pass filter to the image of Sasuke of the same $\sigma$ as shown in Figure 3.2b.

A high value of $\sigma$ is necessary to form a blur image and similarly a high value of $\sigma$ leads to the capturing of more prominent part of the image which is to be passed through the high pass filter. We applied trial and error for tweaking the values of $\sigma$ and the %age of first image to be blended with the second one. After a few tries we found that $\sigma$ of 30 gave the best result and 0.65 of the first image added to the remaining of 0.35 of the second image as shown in Figure 3.3.

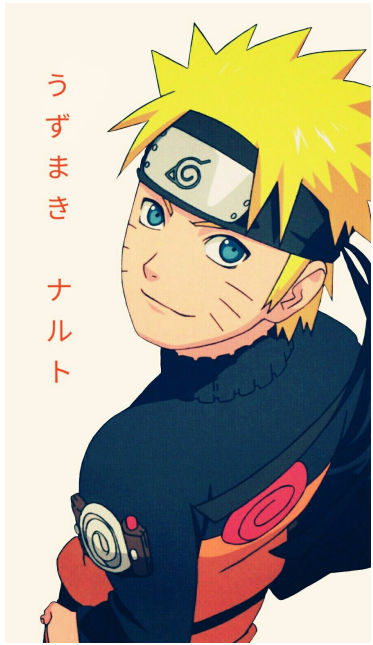

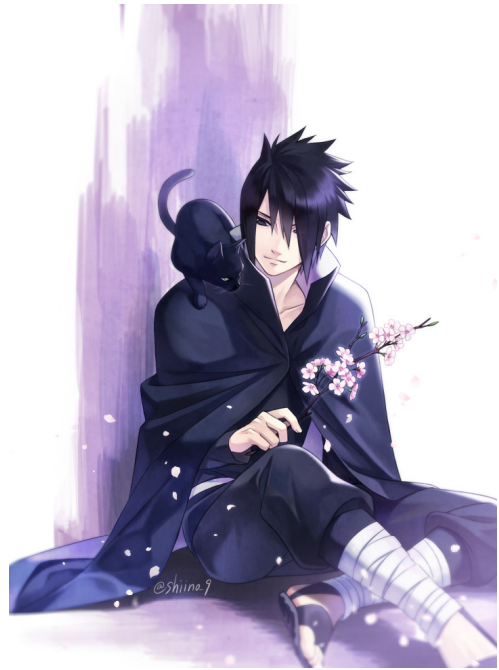

(a) The original image of Naruto (Source:pinimg.com)

(b) The original image of Sasuke. (Source:cdninstagram.com)

Figure 3.1: A figure containing the high resolution images of Naruto and Sasuke.

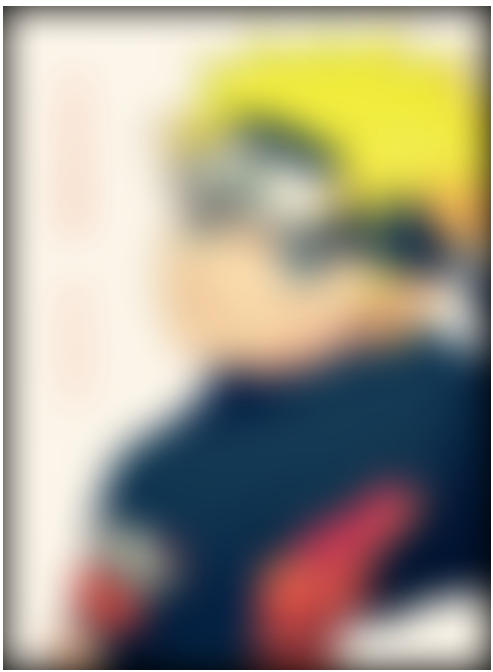

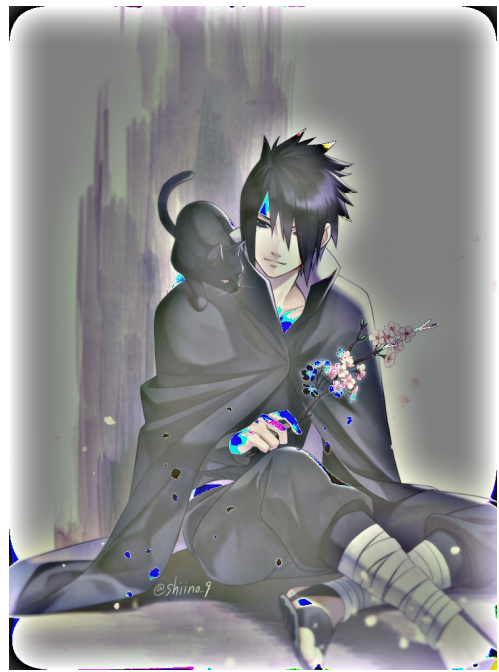

(a) The image of Naruto got by applying a low pass filter of $\sigma = 30$

(b) The image of Sasuke got by applying a high pass filter of $\sigma = 30$

Figure 3.2: A figure containing the high resolution filtered images of Naruto and Sasuke.

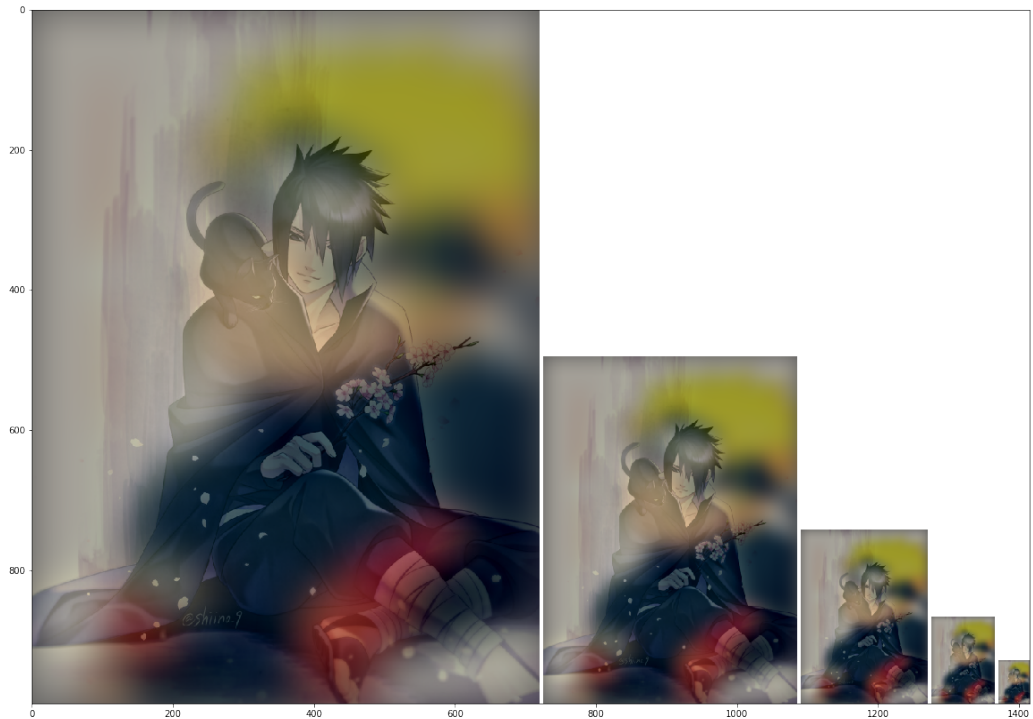

Figure 3.3: The blended image for Naruto (low pass) and Sasuke (high pass) with $\sigma = 30$.

## 3.2 Applying to images of Olive and Palm tree

We load the image of the olive tree first and then the image of palm tree as shown in Figure 3.4. The first image of olive is passed through a lowpass filter (as shown in Figure 3.5a) and the second image of palm is passed through a high pass filter (as shown in Figure 3.5b).

We tweaked the value of the $\sigma$ to be 10, manually after trail and errors for best results. We then blended $0.5\times$ the first image with $0.5 \times$ the second image to get a pure hybrid image as shown in Figure 3.6. So, now since the image of Olive tree is passed through a low pass filter, it will

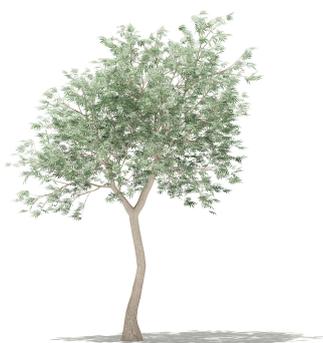
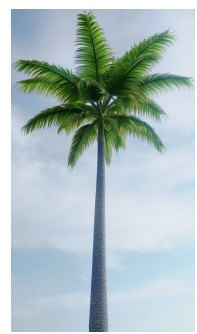

(b) The original image of the palm tree
(a) The original image of the olive tree(Source:Google)
(Source:digitaloceanspaces.com)

Figure 3.4: A figure containing the high resolution images of Olive and Palm trees.

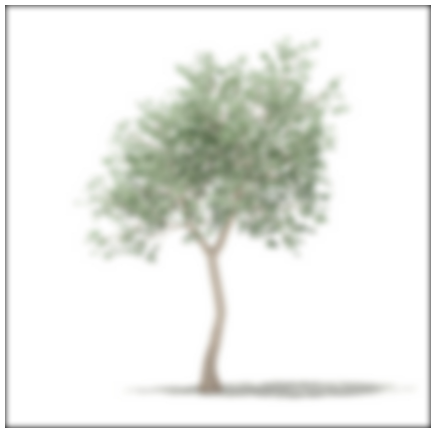
(a) The image of the olive tree (low pass)

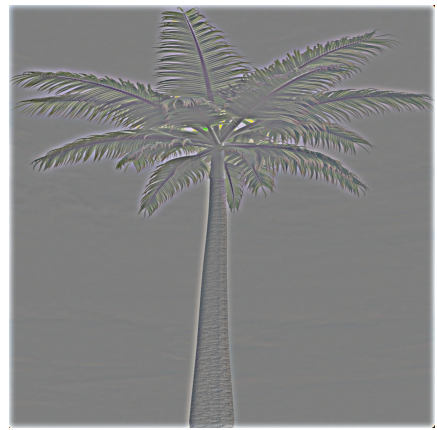
(b) The image of the palm tree (high pass)

Figure 3.5: The same sized version of olive and palm trees by passing a cutoff frequency $\sigma = 10$.

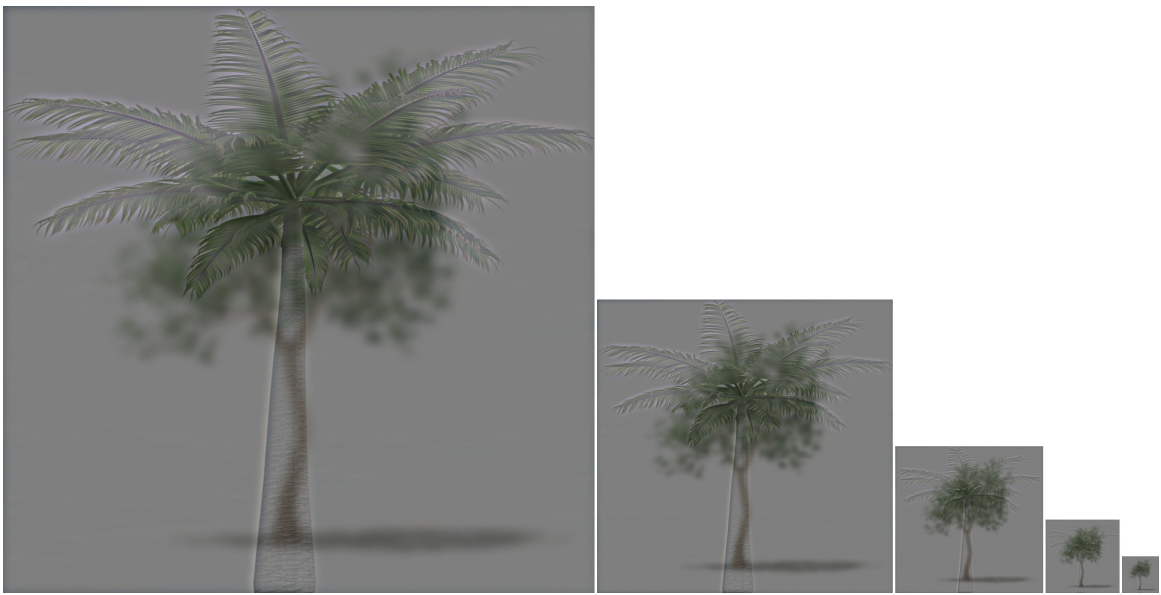

Figure 3.6: The blended image for olive (low pass) and palm tree (high pass) with $\sigma = 10$.

be visible clearly from a distance, and simultaneously if we look closely, we will see the image of the palm tree to be prominent.

## 3.3 Applying to images of Swami Vivekananda and Ramakrishna

We load the original image (Figure 3.7a) of Swami Vivekananda and apply a low pass filter to it as shown in Figure 3.8a. We do the same for the original image (Figure 3.7b) of Ramakrishna but instead apply a high pass filter to it as shown in Figure 3.8b.

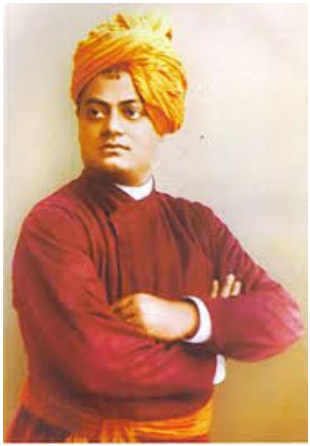
(a) The original image of Swami Vivekananda

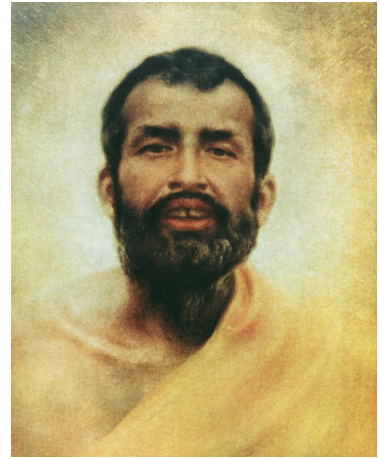
(b) The original image of Ramakrishna

Figure 3.7: The original images of Swami Vivekananda and Ramakrishna.

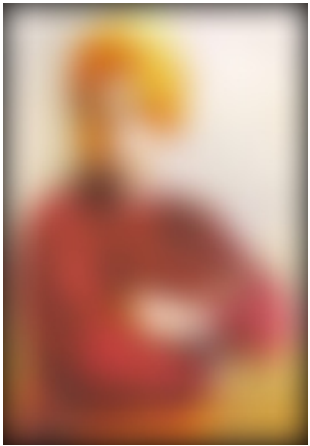
(a) The image of Swami Vivekananda (low pass)

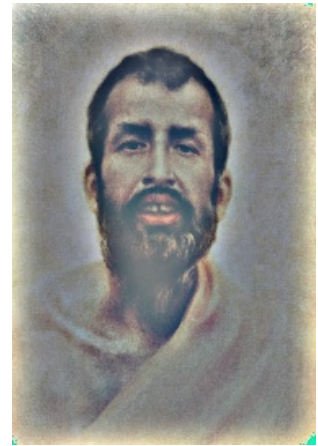
(b) The image of Ramakrishna (high pass)

Figure 3.8: The images of Swami Vivekananda and Ramakrishna by applying a filter of $\sigma = 10$.

We do a lot of trail and error for the amount of blending of the first image to be added to the second image. After a lot of trail and error we find that the $0.85 \times$ the first image is added with $0.15 \times$ the second image then it gives a good result as shown in Figure 3.9.

When we used a filter size of 20, we got a very blurry image of Swami Vivekananda and a very prominent image for Ramakrishna. This is because the color of the two images are in contrast to each other. The high pass image is very prominent because the black color of Ramakrishna dominates the blurry image when viewed from even a distance. We carefully tweaked and increased the concentration of the first image (i.e., the blurry image of Swami Vivekananda) and found that after adding $0.85\times$ the first image, it is quite visible clearly from a far distance as shown in the low pass image.

From the above unique example, we made it clear that we even need to manually tweak the amount of weight the first image needs to be given in the blended version of the image. It may be that the second image needs a high value of $\sigma$ to capture the surroundings of the image to get the high pass value of the image, and the first image may get too blurry to be visible even from a distance. This means there is a tradeoff between the cutoff frequency and the amount of weight that needs to be given to the first image. We will discuss more about these limitations in the next section.

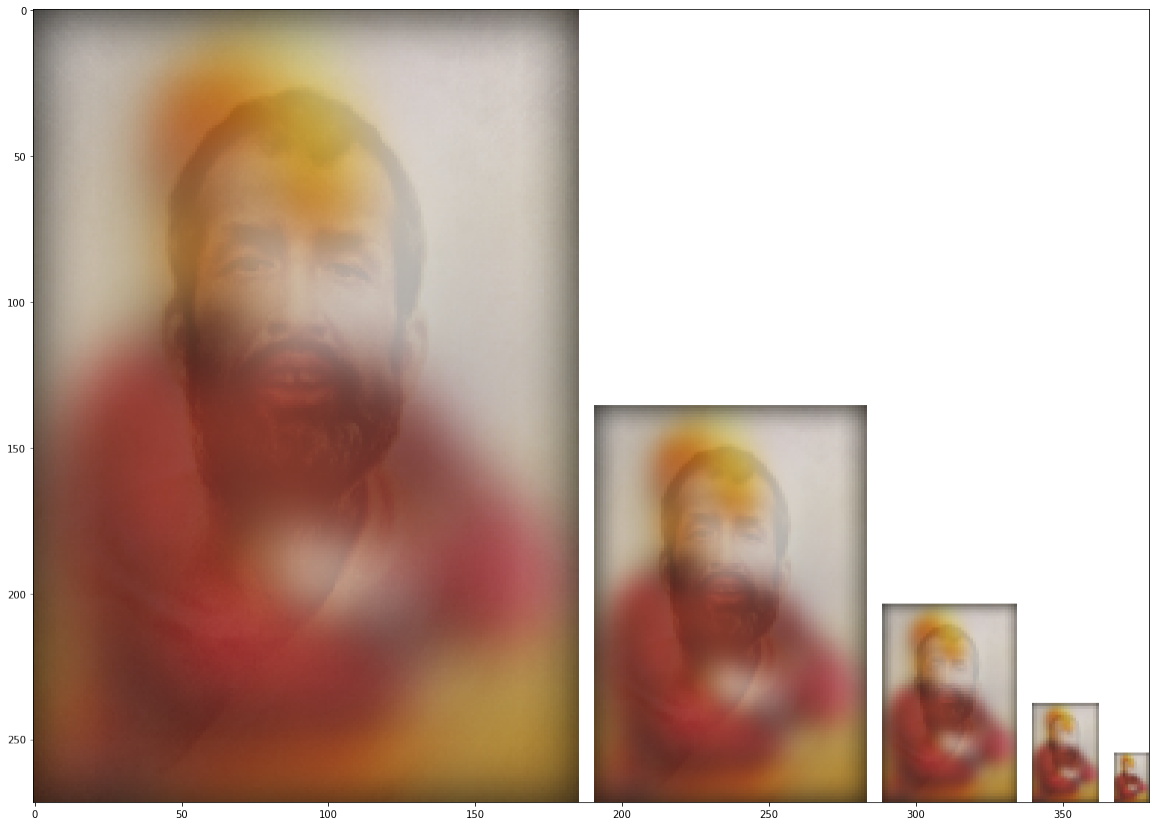

Figure 3.9: The blended image for Vivekananda (low pass) and Ramakrishna (high pass) with $\sigma = 10$.

# Chapter 4

# Conclusion

We have tested the algorithm in the following system as shown in Figure 4.1. We have created an additional script to measure the size $Vs$ time needed for each image and plot the details for analysis. We have found that we need to tweak the parameters like selection of $\sigma$ and the amount of weight the first image needs to be given in the blending of hybrid image. We found that sometimes it needs brute force analysis to check which $\sigma$ and values of weights are suitable for getting the best results in the blending of images.

## 4.1 An analysis of Size and Computation time

We have created a automation script which took almost 15 hours to run on the above machine with the specifications as shown in Figure 4.1. The script took a array of filters of sizes 2,4,5,7,10,15,20,25 and 30. The scatter plot for the filters and the size of the images (i.e., $size = height \times width \times no.of channels$) is shown in fig. 4.2 and fig. 4.3. It recorded the time required for each of the 27 images that were given to it, by applying a highpass as well as a low pass filter. It was found that the low pass filter took a significant amount of time than the low pass filter computation. We doubt that this may be due to the fact that the images that were passed to the lowpass filter were cached and so it took some less amount of time for the high pass filter of same kernel size. It may also be for different reason of numpy optimization that this resulted.

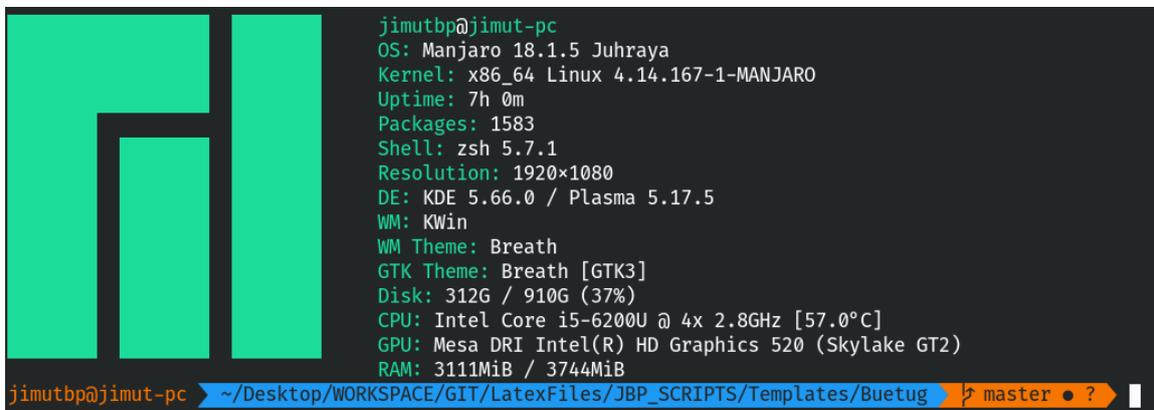

Figure 4.1: The system specifications for the computer in which the algorithm was tested.

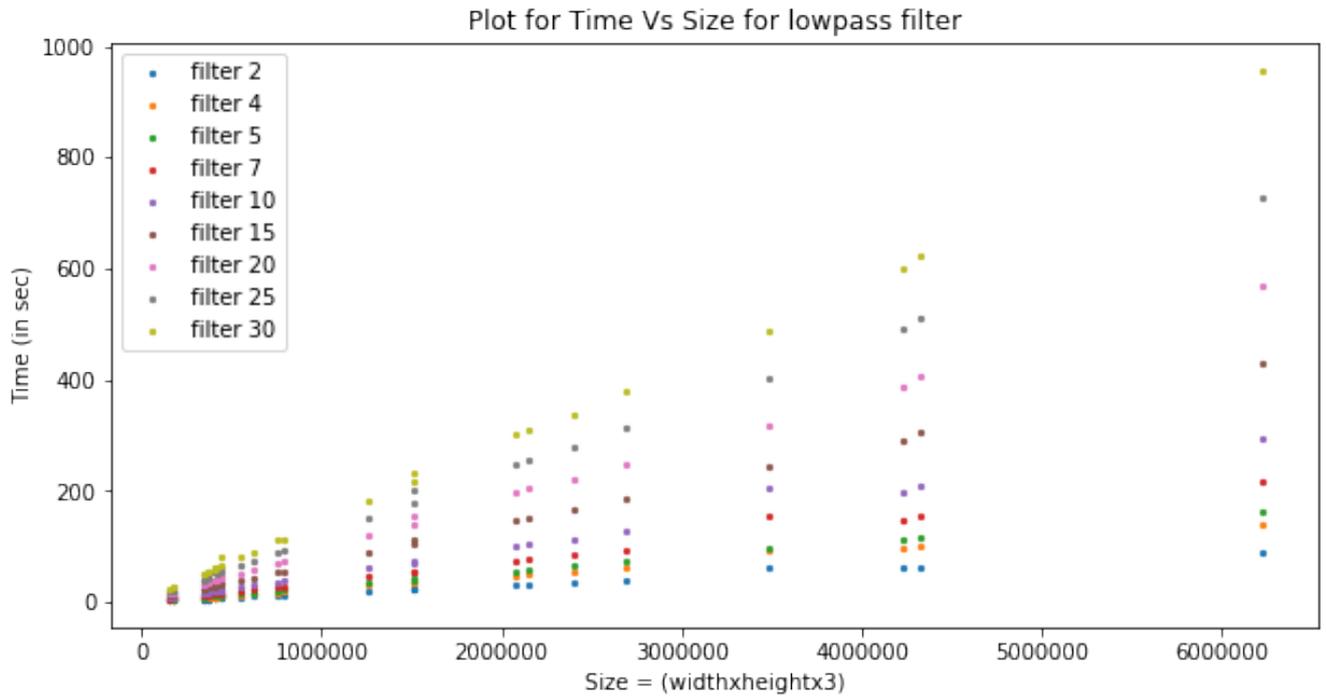

Figure 4.2: The graph of Time (Y-axis) Vs Size (X-axis) for low pass filters.

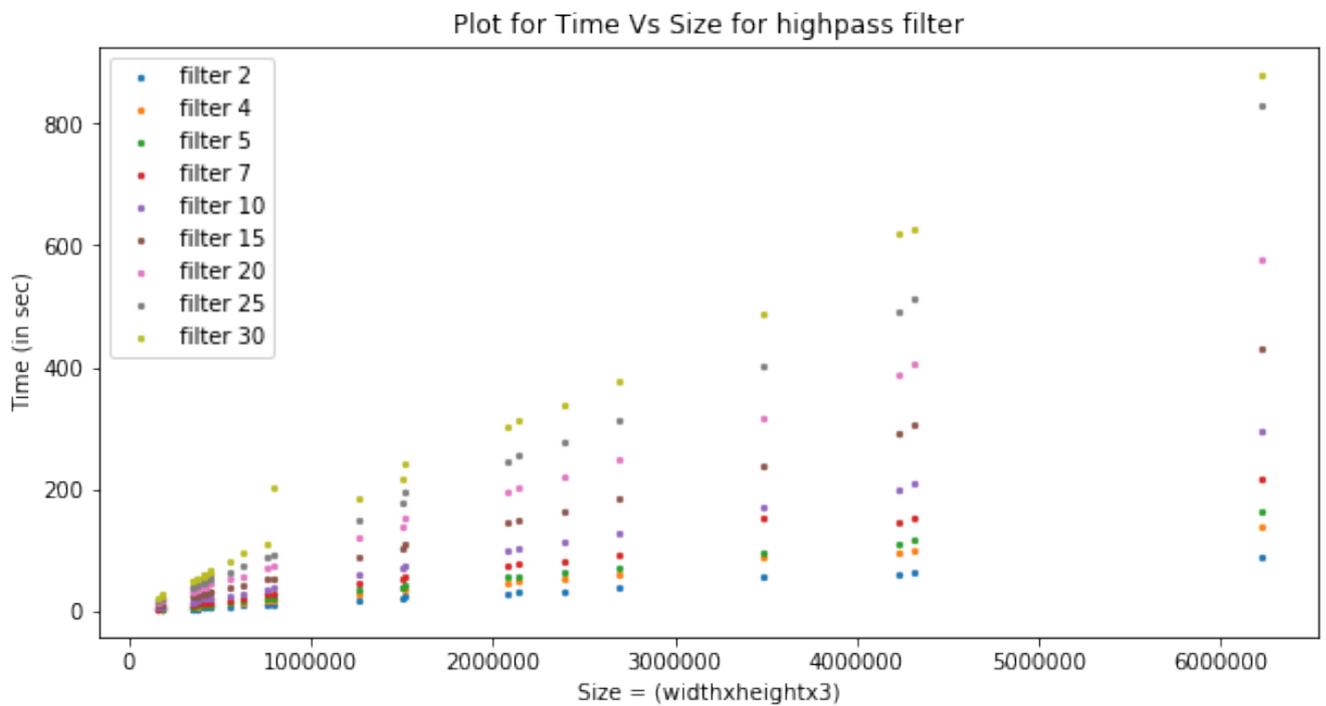

Figure 4.3: The graph of Time (Y-axis) Vs Size (X-axis) for high pass filters.

## 4.2 Can this be applied to a real time system?

**No!** certain images of dimension about 1500*1500 took about 1000 seconds to just blur using the lowpass kernel of size $\sigma = 30$. So, it is almost impossible for making it a real time system. Also the more the size of the image, the more will be the value of $\sigma$ to get good results, so for this reason, it is necessary to make faster methods which will outperform this and make a real time system, perhaps by selecting certain regions of images rather than computing on the whole image. This method also has a limitation that we might need to choose the value of weights that needed to be blended with the other image manually, so we need some intelligent methods which can overcome this limitation.

# Appendix A

# Codes

## A.1 Python code corresponding to the filter2D and hybrid image formation

Here we build our own version of filter2D function of opencv by using numpy.

```python
import numpy as np
#### DO NOT IMPORT cv2

def my_imfilter(image, filter):
    """
    Apply a filter to an image. Return the filtered image.

    Args
    - image: numpy nd-array of dim (m, n, c)
    - filter: numpy nd-array of dim (k, k)
    Returns
    - filtered_image: numpy nd-array of dim (m, n, c)

    HINTS:
    - You may not use any libraries that do the work for you. Using
      numpy to work
      with matrices is fine and encouraged. Using opencv or similar to
      do the
      filtering for you is not allowed.
    - I encourage you to try implementing this naively first, just be
      aware that
```

```python
     it may take an absurdly long time to run. You will need to get a
      function
     that takes a reasonable amount of time to run so that I can verify
     your code works.
     - Remember these are RGB images, accounting for the final image
      dimension.
     """

     assert filter.shape[0] % 2 == 1
     assert filter.shape[1] % 2 == 1

     ############################
     ### TODO: YOUR CODE HERE ###

     # works for any kind of image, whether grayscale or 3-channel RGB
      or any channel image
     filtered_image = image.copy()                                  # copying
      the filtered image
     image_dimension=image.shape                                    # capturing
      the shape of the image
     filter_dimension=filter.shape                                  # capturing
      the shape of the filter
     height=image_dimension[0]                                      # x
     width=image_dimension[1]                                       # y
     flt_dim1=filter_dimension[0]                                   # x    works
      for non-square filters
     flt_dim2=filter_dimension[1]                                   # y    works
      for non-square filters
     pad_height=int((flt_dim1-1)/2)                                 # estimating
       the size of the padding of height with the help of filter
     pad_width=int((flt_dim2-1)/2)                                  # estimating
       the size of the padding of width with the help of filter
     pad_mat=np.zeros((height+2*pad_height,width+2*pad_width,3))
     # this part of code creates the image inside the padded version of
      the numpy zero matrix
     pad_mat[pad_height: height + pad_height, pad_width: width +
      pad_width] = image
     for d in range(len(image[0][0])):                              # going
      through all the channels of the images in the case if it is not in
       grayscale
```

```python
45        for i in range(len(image)):                       # x value iteration or height
46            for j in range(len(image[0])):                # y value or width
47                # now using the padded 3D numpy array or 1D array as the case may be, we apply gaussian blur.
48                filtered_image[i][j][d] = sum(sum(np.multiply(filter, pad_mat[i:i+flt_dim1,j:j+flt_dim2,d])))
49
50    ### END OF STUDENT CODE ####
51    ############################
52
53    return filtered_image
54
55 def create_hybrid_image(image1, image2, filter,first_weight, filter_type):
56    """
57    Takes two images and creates a hybrid image. Returns the low
58    frequency content of image1, the high frequency content of
59    image 2, and the hybrid image.
60
61    Filter type of 1 generates the inverse of a Gaussian filter with the central point
62    at the inverse peak. The second type of filter generates the (N*N) sobel filter.
63
64    Args
65    - image1: numpy nd-array of dim (m, n, c)
66    - image2: numpy nd-array of dim (m, n, c)
67    Returns
68    - low_frequencies: numpy nd-array of dim (m, n, c)
69    - high_frequencies: numpy nd-array of dim (m, n, c)
70    - hybrid_image: numpy nd-array of dim (m, n, c)
71
72    HINTS:
73    - You will use your my_imfilter function in this function.
74    - You can get just the high frequency content of an image by removing its low
75      frequency content. Think about how to do this in mathematical terms.
```

```python
    - Don't forget to make sure the pixel values are >= 0 and <= 1.
     This is known
      as 'clipping'.
    - If you want to use images with different dimensions, you should
     resize them
      in the notebook code.
    """

    assert image1.shape[0] == image2.shape[0]
    assert image1.shape[1] == image2.shape[1]
    assert image1.shape[2] == image2.shape[2]

    ############################
    # low pass filter is the normal gaussian distribution

    low_frequencies = my_imfilter(image1,filter)
    if filter_type == 1:
      # high pass filter is the negative of the gaussian distribution
      h=-filter
      h[int(filter.shape[0]/2),int(filter.shape[0]/2)] = filter[int(filter.shape[0]/2),int(filter.shape[0]/2)]
      h[int(filter.shape[0]/2),int(filter.shape[0]/2)] = h[int(filter.shape[0]/2),int(filter.shape[0]/2)] -h.sum()
      high_frequencies = my_imfilter(image2,h)
      hybrid_image = first_weight*low_frequencies + (1-first_weight)*high_frequencies
      hybrid_image = np.clip(hybrid_image, 0.0, 1.0)
    elif filter_type == 2:
      high_frequencies =  image2 - my_imfilter(image2, filter)
      hybrid_image = first_weight*low_frequencies + (1-first_weight)*high_frequencies
      #print(hybrid_image.max())
      hybrid_image = np.clip(hybrid_image, 0.0, 1.0)
      #print(hybrid_image.max())
    ### END OF STUDENT CODE ####
    ############################

    return low_frequencies, high_frequencies, hybrid_image
```

## A.2 Python code corresponding to the collection of data for time size analysis

This is a script which automatically finds the time and size for each of the images by applying a variety of kernel sizes including sizes of 2,4,5,7,10,15,20,25 and 30. The time, size of the image and kernel type data is recorded in a json file and is later used for analysis.

```python
"""
Auto-script for collecting data for higher dimensional visualization.

This script takes all the images and computes the time taken by
    lowpass and highpass filter for
every images and stores it in a json file.

Author: Jimut Bahan Pal
Dated: 27th January, 2019
"""

import os
import cv2
import sys
import json
import time
import matplotlib
import numpy as np
from tqdm import tqdm
import matplotlib.pyplot as plt
from utils import vis_hybrid_image, load_image, save_image
from student_code import my_imfilter, create_hybrid_image

with open('json_data.json') as f:
    data = json.load(f)
print(data)

pictures_mydata = ['cowboyhat.jpg','lambo.jpg','mexican_house.jpg','
    palm.jpg','sasuke.png','viveka.jpg',
'ferrari.jpg','mattd.jpg','naruto.jpg','ramakrishna.jpg','tom.jpg','
    ylambo.jpg',
```

```python
'housebung.jpg','mexican_hat.jpg','olive.jpg','sasuke.jpg','tomsad.
    jpg']

given_data=['bicycle.bmp', 'cat.bmp', 'einstein.bmp', 'marilyn.bmp
    ','submarine.bmp',
'bird.bmp','dog.bmp', 'fish.bmp', 'motorcycle.bmp', 'plane.bmp']

cutoff_freq = [2,4,5,7,10,15,20,25,30]

path_ = "../data/mydata/"
full_data = data.copy()

for pictures in tqdm(pictures_mydata):
    full_path = path_+pictures
    try:
        if len(data[pictures])>0:
            print("Skipping ",pictures)
            continue
    except:
        print("Continuing...")
    picture_data = {}
    image1 = load_image(full_path)
    print(pictures," :: loaded ")
    picture_data["height"] = image1.shape[0]
    picture_data["width"] = image1.shape[1]
    for freq in tqdm(cutoff_freq):
        filter = cv2.getGaussianKernel(ksize=freq*4+1,
                                sigma=freq)
        filter = np.dot(filter, filter.T)
        start = time.time()
        print(" Applying low pass filter on ",pictures)
        blurry_dog = my_imfilter(image1, filter)
        done = time.time()
        elapsed_lowpass = done - start
        print("Time for low pass filter of size ",freq," : ",
    elapsed_lowpass)
        print(" Applying high pass filter on ",pictures)
        h=-filter
        h[int(filter.shape[0]/2),int(filter.shape[0]/2)] = filter[int
    (filter.shape[0]/2),int(filter.shape[0]/2)]
```

```
67            h[int(filter.shape[0]/2),int(filter.shape[0]/2)] = h[int(
       filter.shape[0]/2),int(filter.shape[0]/2)] -h.sum()
68            start = time.time()
69            high_frequencies = my_imfilter(image1,h)
70            done = time.time()
71            elapsed_highpass = done - start
72            print("Time for high pass filter of size ",freq," : ",
       elapsed_highpass)
73            new_info = {
74                "elapsed_lowpass":elapsed_lowpass,
75                "elapsed_highpass":elapsed_highpass,
76                }
77            print(json.dumps(new_info, indent=4, sort_keys=True))
78            picture_data[freq] = new_info
79        print("Summary of Image ",pictures," :::=> ")
80        #print(json.dumps(picture_data, indent=4, sort_keys=True))
81        full_data[pictures] = picture_data
82        with open('json_data.json', 'w') as f:
83            json.dump(full_data, f,ensure_ascii=False, indent=4)
```

## A.3  Python code for plotting the data for time size analysis

This code takes the collected data generated from the above script and uses that data to plot according to the kernel size, image size and time using different colours.

```
1  """
2  Script to collect the data from the json files and plot accordingly.
       This plots the time Vs size of images along with the size of
       kernels.
3
4  Author: Jimut Bahan Pal
5  Date: 27th January 2019
6  """
7
8
9  import json
10 import matplotlib.pyplot as plt
11
12 # collect the data
13 with open('json_data_given.json') as f:
14     data1 = json.load(f)
```

```python
with open('json_data_myfiles.json') as f:
    data2 = json.load(f)

filter_2 = []
filter_4 = []
filter_5 = []
filter_7 = []
filter_10 = []
filter_15 = []
filter_20 = []
filter_25 = []
filter_30 = []

sizes = []
# put the data in list
for item in data1:
    size = data1[item]["height"]*data1[item]["width"]*3
    sizes.append(size)
    filter_2.append([data1[item]["2"]["elapsed_lowpass"],data1[item]["2"]["elapsed_highpass"]])
    filter_4.append([data1[item]["4"]["elapsed_lowpass"],data1[item]["4"]["elapsed_highpass"]])
    filter_5.append([data1[item]["5"]["elapsed_lowpass"],data1[item]["5"]["elapsed_highpass"]])
    filter_7.append([data1[item]["7"]["elapsed_lowpass"],data1[item]["7"]["elapsed_highpass"]])
    filter_10.append([data1[item]["10"]["elapsed_lowpass"],data1[item]["10"]["elapsed_highpass"]])
    filter_15.append([data1[item]["15"]["elapsed_lowpass"],data1[item]["15"]["elapsed_highpass"]])
    filter_20.append([data1[item]["20"]["elapsed_lowpass"],data1[item]["20"]["elapsed_highpass"]])
    filter_25.append([data1[item]["25"]["elapsed_lowpass"],data1[item]["25"]["elapsed_highpass"]])
    filter_30.append([data1[item]["30"]["elapsed_lowpass"],data1[item]["30"]["elapsed_highpass"]])
for item in data2:
    size = data2[item]["height"]*data2[item]["width"]*3
    sizes.append(size)
    filter_2.append([data2[item]["2"]["elapsed_lowpass"],data2[item]["2"]["elapsed_highpass"]])
```

```python
46        filter_4.append([data2[item]["4"]["elapsed_lowpass"],data2[item][
          "4"]["elapsed_highpass"]])
47        filter_5.append([data2[item]["5"]["elapsed_lowpass"],data2[item][
          "5"]["elapsed_highpass"]])
48        filter_7.append([data2[item]["7"]["elapsed_lowpass"],data2[item][
          "7"]["elapsed_highpass"]])
49        filter_10.append([data2[item]["10"]["elapsed_lowpass"],data2[item
          ]["10"]["elapsed_highpass"]])
50        filter_15.append([data2[item]["15"]["elapsed_lowpass"],data2[item
          ]["15"]["elapsed_highpass"]])
51        filter_20.append([data2[item]["20"]["elapsed_lowpass"],data2[item
          ]["20"]["elapsed_highpass"]])
52        filter_25.append([data2[item]["25"]["elapsed_lowpass"],data2[item
          ]["25"]["elapsed_highpass"]])
53        filter_30.append([data2[item]["30"]["elapsed_lowpass"],data2[item
          ]["30"]["elapsed_highpass"]])
54  fig, ax = plt.subplots(figsize=(10,5))
55  size_plot = 5
56
57  # plot for low pass filter
58  plt.scatter(sizes,[filter_2[0] for filter_2 in filter_2],s=size_plot,
      cmap='#ba1616',label='filter 2')
59  plt.scatter(sizes,[filter_4[0] for filter_4 in filter_4], s=size_plot
      ,cmap='#168bba',label='filter 4')
60  plt.scatter(sizes,[filter_5[0] for filter_5 in filter_5], s=size_plot
      ,cmap='#0cf556',label='filter 5')
61  plt.scatter(sizes,[filter_7[0] for filter_7 in filter_7],s=size_plot,
      cmap='#f50cf2 ',label='filter 7')
62  plt.scatter(sizes,[filter_10[0] for filter_10 in filter_10], s=
      size_plot,cmap='#ddf50c',label='filter 10')
63  plt.scatter(sizes,[filter_15[0] for filter_15 in filter_15], s=
      size_plot,cmap='#f5730c',label='filter 15')
64  plt.scatter(sizes,[filter_20[0] for filter_20 in filter_20], s=
      size_plot,cmap='#760cf5',label='filter 20')
65  plt.scatter(sizes,[filter_25[0] for filter_25 in filter_25], s=
      size_plot,cmap='#070708',label='filter 25')
66  plt.scatter(sizes,[filter_30[0] for filter_30 in filter_30], s=
      size_plot,cmap='#23d58c',label='filter 30')
67
68
69  #ax.axis('equal')
```

```python
70  leg = ax.legend();
71  plt.title('Plot for Time Vs Size for lowpass filter');
72  plt.xlabel('Size = (widthxheightx3)')
73  plt.ylabel('Time (in sec)')
74
75
76
77  # plot for high pass filter
78  fig, ax = plt.subplots(figsize=(10,5))
79  size_plot = 5
80  plt.scatter(sizes,[filter_2[1] for filter_2 in filter_2],s=size_plot,
       cmap='#ba1616',label='filter 2')
81  plt.scatter(sizes,[filter_4[1] for filter_4 in filter_4], s=size_plot
       ,cmap='#168bba',label='filter 4')
82  plt.scatter(sizes,[filter_5[1] for filter_5 in filter_5], s=size_plot
       ,cmap='#0cf556',label='filter 5')
83  plt.scatter(sizes,[filter_7[1] for filter_7 in filter_7],s=size_plot,
       cmap='#f50cf2 ',label='filter 7')
84  plt.scatter(sizes,[filter_10[1] for filter_10 in filter_10], s=
       size_plot,cmap='#ddf50c',label='filter 10')
85  plt.scatter(sizes,[filter_15[1] for filter_15 in filter_15], s=
       size_plot,cmap='#f5730c',label='filter 15')
86  plt.scatter(sizes,[filter_20[1] for filter_20 in filter_20], s=
       size_plot,cmap='#760cf5',label='filter 20')
87  plt.scatter(sizes,[filter_25[1] for filter_25 in filter_25], s=
       size_plot,cmap='#070708',label='filter 25')
88  plt.scatter(sizes,[filter_30[1] for filter_30 in filter_30], s=
       size_plot,cmap='#23d58c',label='filter 30')
89  #ax.axis('equal')
90  leg = ax.legend();
91  plt.title('Plot for Time Vs Size for highpass filter');
92  plt.xlabel('Size = (widthxheightx3)')
93  plt.ylabel('Time (in sec)')
```